\documentclass[letterpaper]{article} 
\usepackage{aaai2026}  
\usepackage{amsthm}
\usepackage{times}  
\usepackage{helvet}  
\usepackage{courier}  
\usepackage[hyphens]{url}  
\usepackage{graphicx} 
\usepackage{natbib}  
\usepackage{caption} 
\usepackage{algorithm}
\usepackage{algorithmic}
\usepackage{amsfonts}

\usepackage{newfloat}
\usepackage{listings}
\DeclareCaptionStyle{ruled}{labelfont=normalfont,labelsep=colon,strut=off} 
\floatstyle{ruled}
\newfloat{listing}{tb}{lst}{}
\floatname{listing}{Listing}
\usepackage[most]{tcolorbox}
\usepackage{caption} 
\newtcolorbox[auto counter]{mybox}[2][]{%
  colback=gray!5,
  colframe=black!75,
  fonttitle=\bfseries,
  title=Box~\thetcbcounter: #2, 
  #1
}

\title{Personalization of Large Foundation Models for Health Interventions}
\author{
    Stefan Konigorski\textsuperscript{\rm 1,2},
    Johannes E. Vedder\textsuperscript{\rm 1},
    Babajide Alamu Owoyele\textsuperscript{\rm 1},
    İbrahim Özkan\textsuperscript{\rm 1}
}
\affiliations{
    \textsuperscript{\rm 1}Hasso Plattner Institute for Digital Engineering, University of Potsdam, Potsdam, Germany\\
    \textsuperscript{\rm 2}Windreich Department of Artificial Intelligence and Human Health, Icahn School of Medicine at Mount Sinai, New York, USA\\

    stefan.konigorski@hpi.de
}

\usepackage{bibentry}

\begin{document}

\maketitle

\begin{abstract}
Large foundation models (LFMs) transform healthcare AI in prevention, diagnostics, and treatment. However, whether LFMs can provide truly personalized treatment recommendations remains an open question. Recent research has revealed multiple challenges for personalization, including the fundamental generalizability paradox: models achieving high accuracy in one clinical study perform at chance level in others, demonstrating that personalization and external validity exist in tension. This exemplifies broader contradictions in AI-driven healthcare: the privacy-performance paradox, scale-specificity paradox, and the automation-empathy paradox. As another challenge, the degree of causal understanding required for personalized recommendations, as opposed to mere predictive capacities of LFMs, remains an open question. N-of-1 trials -- crossover self-experiments and the gold standard for individual causal inference in personalized medicine -- resolve these tensions by providing within-person causal evidence while preserving privacy through local experimentation. Despite their impressive capabilities, this paper argues that LFMs cannot replace N-of-1 trials. We argue that LFMs and N-of-1 trials are complementary: LFMs excel at rapid hypothesis generation from population patterns using multimodal data, while N-of-1 trials excel at causal validation for a given individual. We propose a hybrid framework that combines the strengths of both to enable personalization and navigate the identified paradoxes: LFMs generate ranked intervention candidates with uncertainty estimates, which trigger subsequent N-of-1 trials. Clarifying the boundary between prediction and causation and explicitly addressing the paradoxical tensions are essential for responsible AI integration in personalized medicine. 
\end{abstract}


\section{Introduction}

Large foundation models (LFMs) are large-scale neural networks trained on extensive, heterogeneous data using self-supervised objectives to learn general-purpose representations \cite{vaswani2017attention}. They exhibit strong transferability, enabling adaptation to diverse downstream tasks through fine-tuning or prompting without retraining from scratch. By capturing broad statistical and semantic structure across modalities, they serve as a unifying foundation for modern artificial intelligence systems \cite{bommasani2022opportunitiesrisksfoundationmodels}.

In healthcare, LFMs have initiated a paradigm shift toward data-driven, generalizable AI systems. They have been utilized in various areas \cite{moor2023foundation} with an increasing number of publications in artificial intelligence, machine learning venues, and high-profile medical journals. Some examples include LFMs pretrained on extensive collections of electronic health records \cite[EHR;][]{du2025ehreval, steinberg2024motor}, for medical imaging and pathology \cite{xu2024whole, kondepudi2025foundation, kim2024transparent}, in genomics and other molecular domains \cite{ali2025largelanguagemodels, dalla-torre2024nucleotide, fu2025foundation}, and imaging-text models \cite{kim2024transparent}. 
Such LFMs allow tasks such as disease risk prediction \cite{belyaeva2023multimodalllm}, disease diagnosis \cite{xiang2025vision}, treatment recommendations \cite{alkaeed2025openfoundationmodelshealthcare}, and modeling treatment responses \cite{chen2025unicure}. 
Further examples of LFMs exist across all of healthcare, such as LFMs trained on facial images \cite{haugg2025foundation}, LFMs trained on wearables \cite{erturk2025sensordatafoundationmodels}, and LFMs for biomedical research, such as protein folding \cite{jumper2021highly}.
The application of LFMs in healthcare also yields specific challenges including data privacy, bias, hallucinations, interpretability, high computational costs, and alignment with ethical and regulatory standards.

This paper focuses on using LFMs to personalize health interventions, including medical treatments, lifestyle interventions, and any behavior to maintain or improve health. Examples include what an individual should do to improve their headache or which medication at which dosage is optimal to control high blood pressure. This yields two critical challenges for LFMs: \textit{causal reasoning} and \textit{personalization}. While LFMs are proficient at identifying statistical patterns across large populations, they may lack the counterfactual evidence necessary to determine causal treatment effects for individuals. This raises the question: \textbf{\textit{How can LFMs trained on population data be truly personalized and enable causally-supported recommendations at the individual level?}}

\section{Personalization of LFMs for Health Interventions}

\subsection{Assumptions for Personalization}

It is possible to define general conditions under which population-level LFMs can prescribe optimal treatment recommendations for a given patient. If any of the three conditions in Box~\ref{box:assumptions} are satisfied, optimal personalized treatment suggestions can be obtained. Otherwise, recommendations from LFMs are not guaranteed to provide optimal treatment and may be strongly biased or yield even adverse health outcomes. 
Condition 1 is satisfied only in scarce situations. Conditions 2 and 3 apply to a broader range of medical and behavioral situations, but require a detailed understanding of the context, a well-defined model that may or may not require correct knowledge of the causal structure of treatment and health outcomes, and sufficient information about the personal characteristics and context of the patient, which may be further complicated if treatment effects are time-varying. Another challenge is presented by rare diseases, where, by definition, little or no data is available for learning population-level LFMs.

\begin{mybox}[label={box:assumptions}]{Assumptions for Personalization}
Sufficient conditions for personalized treatment suggestions are any of the following:
\begin{enumerate}
    \item The treatment works for everyone all the time.
    \item Sufficient information is available from the context, and the medication always works in this situation for everyone.
    \item Sufficient information is available from the context and the patient, and the learned model is complex enough to have learned which intervention is optimal for every set of patient characteristics and every context.
\end{enumerate}
\end{mybox}

\subsection{Contributions}

The conditions create additional challenges. Recent research revealed a fundamental \textit{generalizability paradox}: models that achieve high accuracy in one clinical study perform at chance level in others, demonstrating that personalization and external validity are in tension. This exemplifies broader contradictions in AI-driven healthcare, which we characterize as: the \textit{privacy-performance paradox} (personalization requires comprehensive data, yet privacy demands data minimization), the \textit{scale-specificity paradox} (foundation models need massive populations but must serve individuals), and the \textit{automation-empathy paradox} (AI efficiency risks dehumanizing the patient-centered care it aims to enhance).

In the medical and biostatistics literature, the proposed gold standard for obtaining personalized evidence to support treatment suggestions is an N-of-1 trial \cite{nikles2015essential}. In N-of-1 trials, individuals apply the different treatments of interest in crossover periods and collect health outcomes of interest via self-reporting or passively using wearables. Then, statistical analyses and causal inference can be performed on the resulting single-person time series \cite{piccininni2024causal, daza2018causal, konigorski2024digital}. As an advantage, this approach to conducting a new, personalized self-experiment allows to address many of the above-described challenges. Still, questions remain about how to scale the approach and make it available broadly to the population.

We argue that combining LFMs and N-of-1 trials can open new avenues for personalized health interventions. In the following, we first present examples of the current state of personalization in foundation models in healthcare, then describe the challenges and tensions for their application to personalize health interventions in detail, and finally discuss a hybrid framework combining LFMs and N-of-1 trials. This framework navigates the identified paradoxes: it addresses cold-start with population priors while converging on individual evidence; balances scale and specificity through selective validation; and maintains human-centered care through transparent, experimentally grounded personalization.

\subsection{Examples of Existing Personalization in LFMs}

For personalizing health interventions, several LFMs have been proposed that leverage large-scale, multimodal data to tailor medical decisions for individual patients. These models integrate diverse data sources, including EHRs, medical imaging, genomics, and wearables, to enhance the precision of healthcare delivery. Table~\ref{tab:lfm_personalization} summarizes representative approaches that directly include personalization into model building based on existing population-level data or newly to-be-collected individual data, approaches that add finetuning layers for personalization, or also indirect approaches and approaches assuming that no personalization is needed.

\begin{table*}[t]
\centering
\small
\begin{tabular}{|p{0.3cm}|p{3.2cm}|p{2.8cm}|p{3.5cm}|p{4.5cm}|}
\hline
\textbf{No.} & \textbf{Model/Approach} & \textbf{Training Data} & \textbf{Domain} & \textbf{Approach to Personalization} \\
\hline
1 & CausalMed \cite{li2024causalmed} & EHR & Medication recommendations & Causal discovery \& integration of longitudinal patient data in model \\
\hline
2 & HeLM \cite{belyaeva2023multimodalllm} & Clinical features, disease labels, spirometry & Disease risk prediction, treatment recommendations & Personalized recommendations based on group-level characteristics \\
\hline
3 & PH-LLM \cite{Khasentino2025} & Gemini LLM finetuned for text understanding and reasoning & Expert domain knowledge, health recommendations, prediction & Finetune LLM, predict patient-reported outcomes based on measured wearable data \\
\hline
4 & PhysioLLM \cite{fang_physiollm_2024} & None (Provide fitbit data to GPT-4-turbo in prompts) & Rating of achieved personalization & Chat with LLM that has access to individuals' Fitbit summary data \\
\hline
5 & Time2Lang \cite{pillai_time2lang_2025} & Synthetic data of time series with periodicity & Classification tasks in mental health & LLM may be applied to individual's wearables data \\
\hline
6 & Federated fine-tuning \cite{li_open_2025} & None (review paper) & Different biomedical applications & Proposal that federated LFMs enable personalized model tuning \\
\hline
7 & MedAgentSim \cite{almansoori_self-evolving_2025} & None (User prompts in evaluation) & Patient-doctor conversations & Indirectly by creating agent-based simulations considering personal factors \\
\hline
8 & Language-Assisted Medication Recommendation \cite{zhao_addressing_2025} & Finetune existing LLMs on EHR data & Prescription recommendations, personalization not primary aim & Ignore personal characteristics in LLM prompt but include finetuned overall drug-disease relationships \\
\hline
9 & UniCure \cite{chen2025unicure} & Integrates omics \& chemical LFMs & Cancer treatment prediction & Obtain personalized drug ranking vector based on predicted transcriptomic perturbations  \\
\hline
\end{tabular}
\caption{Summary of existing personalization approaches in large foundation models for healthcare. Models leverage diverse data sources and employ various techniques, including causal inference, multimodal integration, federated learning, and real-time sensor analysis, to tailor interventions to individual patients.}
\label{tab:lfm_personalization}
\end{table*}

\subsection{Challenges for Personalizing LFMs}

As described in Table~\ref{tab:lfm_personalization}, several proposed models may yield personalized health recommendations. Still, there are several challenges and limitations to the current models, in addition to the general assumptions that must be satisfied (see Box~\ref{box:assumptions}).

\subsubsection*{Data requirements and cold start.}
Personalization requires sufficient within-person data to capture individual heterogeneity. Most deployments start with too little signal, producing unstable estimates and weak calibration. When a new user has no history, the cold start limits early utility and delays benefit. Longitudinal wearable and self-report data may introduce missing data points, artifacts, and non-wear bias, which can degrade modeling unless explicitly handled \cite{dlacm_cold_start}.

\subsubsection*{Privacy and security.}
Personalized systems aggregate sensitive multimodal traces, including EHR, text, imaging, genomics, and wearable data. Reidentification risk persists even after deidentification, and data custody across the development pipeline remains unclear. Continuously learning models challenge existing compliance frameworks such as HIPAA, GDPR, and FDA post-market controls, especially when audio-visual data is recorded \cite{price2019privacy}.

\subsubsection*{Bias and fairness.}
Foundation models can inherit and amplify existing inequities. Proxies like healthcare costs used as indicators of need can reproduce systemic disparities. Underrepresentation of specific demographic groups leads to uneven error rates and unequal access. Vision-language models in medicine exhibit larger fairness gaps than human experts and can infer protected attributes from images. Many released high-performing benchmark models and tools still lack diverse multi-site validation \cite{bommasani2022opportunitiesrisksfoundationmodels}.

\subsubsection*{Hallucination and reliability.}
Large language models can produce fluent but false outputs, lack awareness of recent medical evidence, and misjudge their own uncertainty. Such behavior is unsafe in clinical contexts without strong verification, retrieval, and safeguard mechanisms \cite{moor2023foundation}.

\section{Four Structural Tensions for Personalization in LFMs Requiring Hybrid Approaches}

We now identify four inherent paradoxical tensions in AI-driven personalization that arise from fundamental mismatches between population-trained models and individual causal inference. These tensions motivate hybrid personalization systems guided by design principles/requirements for integrating foundation models with experimental validation in healthcare.

\subsection{Tension 1: Personalization Versus External Validity}

\textbf{The tension.} Optimizing models for personalized predictions in a specific context inherently reduces their ability to generalize to new contexts, reflecting a structural limitation rather than a calibration issue.

\textbf{Empirical evidence.} \cite{chekroud2024illusory} trained ML models to predict schizophrenia treatment outcomes using trial data. Within-trial performance was strong (Area under the Curve (AUC) $>$ 0.70), but collapsed to chance (AUC $\approx$ 0.50) when applied to independent trials with identical treatments and similar populations.

\textbf{Why this occurs.} Models estimate average effects but cannot determine which subgroup individuals belong to without individual-level data. They also overfit to context-specific idiosyncrasies that don't transfer \cite{subbaswamy2019preventing, hernan2020causal}.

\textbf{Design requirement:} \textit{Systems must quantify prediction uncertainty and trigger experimental validation when uncertainty thresholds (on treatment effect estimates or estimates of differences in treatment effects) are exceeded.}

\subsection{Tension 2: Data Requirements Versus Privacy Protection}

\textbf{The tension.} Effective personalization requires comprehensive individual data, while privacy protection requires data minimization \cite{price2019privacy}. This creates a circular dependency: users won't share data without trust, but systems can't demonstrate trustworthiness without data.

\textbf{Why technical solutions are insufficient.} Differential privacy degrades accuracy \cite{abadi2016deep}, federated learning leaks information through gradients \cite{nasr2019comprehensive}, and multimodal health data resists de-identification, genomic data is inherently identifiable \cite{gymrek2013identifying}, and behavioral patterns create unique fingerprints \cite{de2013unique}. Deidentification approaches are viable solutions for sharing experimental recordings within the treatment ecosystem using masking approaches \cite{owoyele_maskanyone_2024}.

\textbf{Design requirement:} \textit{Individual validation must occur locally on user devices with minimal data transmission. Only with consent should anonymized results contribute to collective learning.}

\subsection{Tension 3: Population-Scale Training Versus Individual Application}

\textbf{The tension.} Foundation models require massive populations for training \cite{moor2023foundation}, but clinical decisions target individuals. Population estimates do not predict individual responses well when heterogeneity is substantial \cite{kravitz_evidence-based_2004}.

\textit{Epistemically}, the "average patient" is a mathematical abstraction. Models trained on millions cannot identify which subgroup any individual belongs to without individual-level evidence. \textit{Economically}, as treatments become targeted, development costs become prohibitive when amortized across smaller populations \cite{kimmelman_paradox_2018}.

\textbf{Design requirement:} \textit{Leverage multimodal data for selective validation \cite{Schneider2023arXiv2309.06455, Fu2023arXiv2302.07547}, reserve experiments for high-stakes or high-uncertainty scenarios where population knowledge is insufficient.}

\subsection{Tension 4: Algorithmic Efficiency Versus Human-Centered Care}

\textbf{The tension.} AI promises efficiency through automation, but medicine involves listening, understanding values, building trust, and responding to suffering with compassion \cite{pot_not_2021}. Algorithmic decision-making risks dehumanizing care by treating individuals as data points \cite{morrow_artificial_2023}.

\textbf{Manifestations.} Foundation models struggle with the narrative and existential dimensions of illness. Black-box opacity prevents meaningful explanation, undermining shared decision-making \cite{kocaballi_personalization_2019,mahesh_advancing_2024}. If AI provides diagnoses, the therapeutic value of clinical encounters may diminish.

\textbf{Design requirement:} \textit{Experimental validation provides interpretable evidence that patients and clinicians can understand together. Patients become active participants in knowledge generation, preserving agency and clinical accountability \cite{konigorski_studyu_2022}.}

\subsubsection{Summary: Why Hybrid Approaches Are Necessary}

These tensions are structural features of applying population-trained models to individual causation. Table~\ref{tab:tensions} summarizes how hybrid frameworks address each.
Hybrid systems integrate population-based hypothesis generation (foundation models' strength) with individual-level experimental validation (providing causal guarantees observational learning cannot achieve).

\begin{table}[h]
\centering
\small
\begin{tabular}{p{0.001\textwidth}p{0.12\textwidth}p{0.135\textwidth}p{0.15\textwidth}}
\hline
 & \textbf{Tension} & \textbf{Why It Occurs} & \textbf{Hybrid Solution} \\
\hline
1 & Personalization vs. external validity & Population patterns don't predict individual responses & LFM generates/speculates hypothesis; N-of-1 trial validates when uncertainty is high \\
\hline
2 & Data requirements vs. privacy protection & Comprehensive data is needed but creates risks & Local experimentation requires minimal data sharing \\
\hline
3 & Population training vs. individual application & Average effects don't determine individual responses & Selective validation for high-stakes/uncertainty cases \\
\hline
4 & Efficiency vs. human-centered care & Algorithms lack narrative dimensions & Experimental evidence is interpretable; patients are active participants \\
\hline
\end{tabular}
\caption{Four structural tensions and hybrid solutions.}
\label{tab:tensions}
\end{table}

\section{Personalizing LFMs with N-of-1 Trials}

\subsection{N-of-1 Trials}

\begin{figure}[t]
\includegraphics[width=1.0\columnwidth]{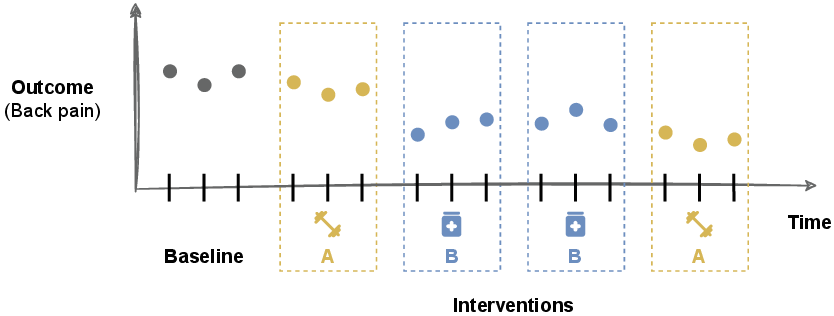}
\caption{Illustration of an N-of-1 trial design with alternating interventions (A and B) following a baseline period to evaluate individual treatment responses over time.}
\label{Fig-Nof1-Design}
\end{figure}

N-of-1 trials are the gold standard for determining which interventions are effective for a specific individual. Unlike traditional studies that estimate average effects across populations, N-of-1 trials focus on a single person. As illustrated in Figure \ref{Fig-Nof1-Design}, the individual participates in repeated, controlled intervention periods that alternate between different treatments or conditions, with outcomes systematically recorded over time. This generates causal evidence about what works best for \textit{that particular person}.
For example, a patient with chronic pain might alternate weekly between two medications over several weeks, rating their pain each day. Statistical analysis of their personal data reveals which medication is more effective for them, regardless of what works "on average" for most patients.

\subsection{Hybrid Framework of personalizing LMFs with N-of-1 trials}

\begin{figure*}[t]
\centering
\includegraphics[width=1.75\columnwidth]{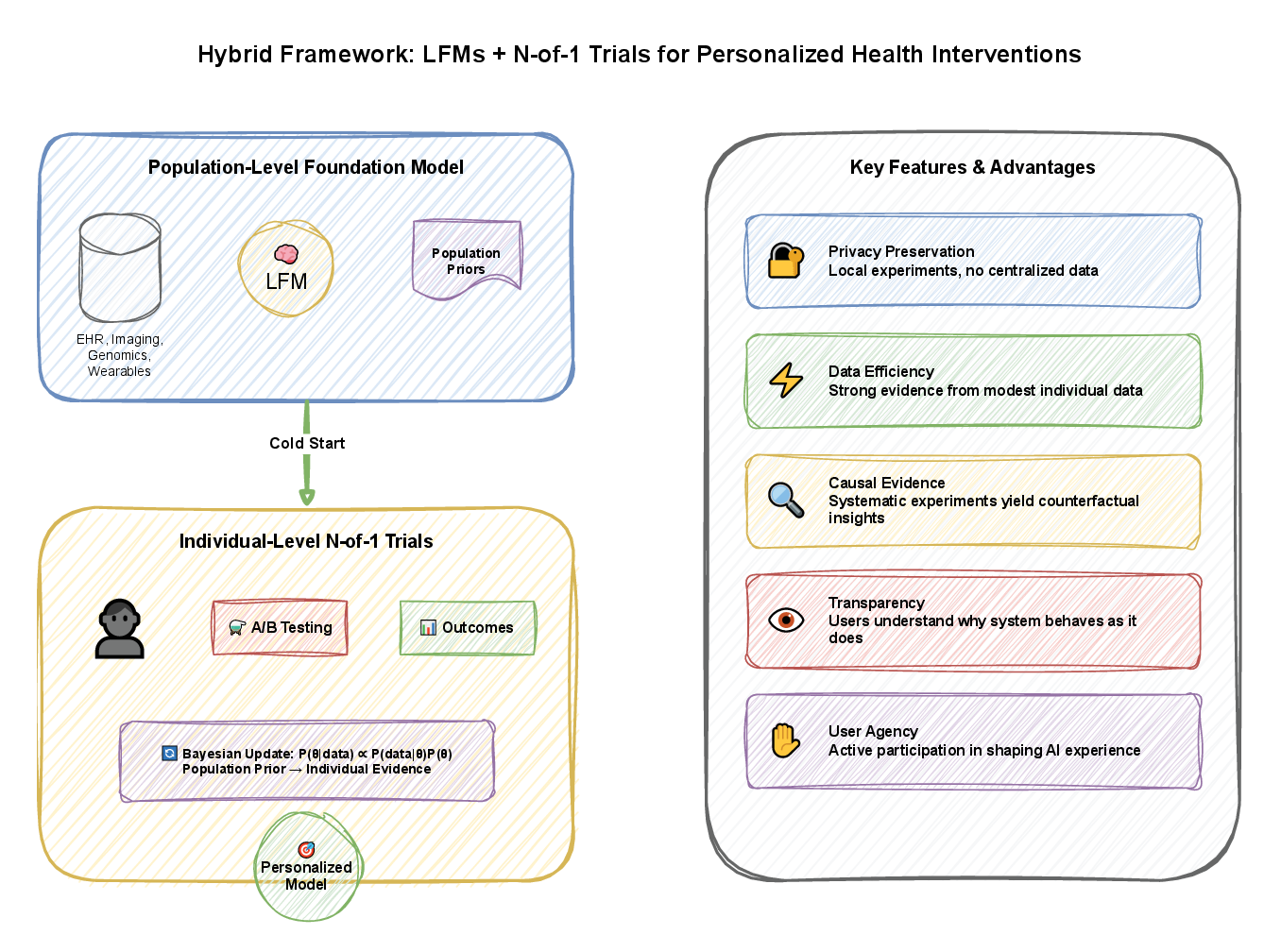} 
\caption{Visual illustration of the hybrid framework combining population-level LFMs with individual-level N-of-1 trials for evidence-based personalization of health interventions.}
\label{Fig-Hybrid-Framework}
\end{figure*}

We propose adapting the N-of-1 trial methodology to personalize foundation models. The core idea is to treat a foundation model as a baseline "digital twin" to build initial hypotheses and treatment suggestions, then dynamically adapts through individualized experiments if needed. Figure \ref{Fig-Hybrid-Framework} illustrates the proposed hybrid framework. We enumerate how that would work in practice below.

\subsubsection{Step 1: Foundation Model as Baseline.}

A large, population-trained foundation model serves as a starting point -- capturing broad patterns and capabilities learned from millions of examples. This represents the "average" knowledge, analogous to population-level treatment effects in medicine.

\subsubsection{Step 2: N-of-1 Trials.}

The system conducts controlled experiments for each user to identify optimal personalization strategies if needed. Rather than assuming what works for most users works for this user, the system tests alternatives directly with that individual. Specifically, the system runs controlled experiments comparing different personalization approaches for the same user:

\begin{itemize}
\item \textbf{Intervention variations:} Compare prompt-tuned version A versus retrieval-augmented generation version B; memory-enabled versus memory-free responses; or formal versus casual communication styles.
\item \textbf{Temporal alternation:} Switch between conditions across sessions or across days, just as clinical trials alternate between treatments.
\item \textbf{Individual analysis:} Measure outcomes -- user satisfaction, task success, engagement -- and analyze results for this specific person, not averaged across all users.
\end{itemize}

The conducted trials might be fixed N-of-1 trials, as shown in Figure \ref{Fig-Nof1-Design}, or adaptive N-of-1 trials \cite{Shrestha2021BayesianBandit, Senarathne2020Bayesian, Meier2023Designing} to compare more than two intervention candidates more efficiently:

\begin{itemize}
\item \textbf{Bayesian integration:} Start with population-level priors (what works for most users), then update with individual evidence. As personal data accumulates, individual patterns increasingly dominate.
\item \textbf{Contextual optimization:} Use methods like contextual bandits to balance exploration (trying new approaches) with exploitation (using what's known to work), tailored to each user's context.
\item \textbf{Uncertainty-aware adaptation:} When confidence in personalization is low, run more experiments. When patterns are clear, apply established preferences consistently.
\end{itemize}

As another way to personalize LFMs in the second step, concepts from N-of-1 trials may inform how user behavior can provide continuous experimental evidence. It might be of interest to incorporate explicit user feedback, such as ratings, corrections, stated preferences, to incorporate implicit signals such as engagement duration, task completion rates, frequency of clarification requests, acceptance of suggestions, or to incorporate more complex behavioral patterns such as time of day preferences, query complexity trends, and interaction styles. These metrics can all help to provide personalized performance indicators, revealing which model configurations serve this individual best.

\subsubsection{Step 3: Dynamic Updating.}

Based on experimental results, the model adapts its behavior for each user, creating a personalized digital twin that combines population knowledge with individual evidence. In further iterative steps, the model should maintain uncertainty awareness and know when to experiment versus when to apply established patterns.

\subsubsection{}

This proposed hybrid approach combines the strengths of population-scale learning (LMFs) with individual-level causal inference (N-of-1 trials), creating personalization that is both effective and trustworthy. It also helps to navigate the described tensions to address key concerns in AI personalization:
\textit{Privacy is preserved} by performing experiments locally on user devices. Personal data need not be centralized or shared. Only if users explicitly consent do anonymized results contribute to collective learning.
The approach is \textit{data efficient} since systematic experimentation yields strong evidence from modest amounts of individual data, avoiding the need for comprehensive personal data collection.
Users understand why the system behaves in specific ways, grounded in their experimental results rather than opaque algorithmic decisions, yielding \textit{transparency}.
Finally, the framework supports \textit{user agency}, since individuals opt into structured experiments, maintain control over their data, and actively participate in shaping their AI experience rather than passively receiving algorithmic prescriptions.

Digital twin frameworks proposed for health interventions in prior work \cite{Qian2021SyncTwin, Holt2024, Sadee2025} may include causal inference and leverage individual-level data to update the model, similar to our proposal. In contrast to our proposal, existing frameworks rely on updating/finetuning these models without generating individual-level experimental evidence of the effectiveness of the trials. The efficacy and differences between approaches may stem from the assumptions outlined in Box~\ref{box:assumptions}.

\section{Illustrative Case Study: Chronic Migraine Management}

To demonstrate the practical application of our hybrid framework, we present an illustrative case study in chronic migraine management -- a condition characterized by high inter-individual variability in treatment response \cite{lipton_migraine_2007}. This case study illustrates how the hybrid framework navigates the identified tensions: the LFM provides efficient hypothesis generation while the N-of-1 trial delivers individual causal evidence, all while preserving privacy and patient agency.

\subsection{Clinical Context}

Consider a patient, Alice, experiencing 12 migraine days per month despite trying multiple preventive medications. Population-level evidence suggests several intervention candidates (e.g., beta-blockers, CGRP inhibitors, lifestyle modifications), yet clinical trials report response rates of only 40--60\%, indicating substantial individual heterogeneity.

\subsection{Step 1: LFM Hypothesis Generation}

An LFM trained on electronic health records, wearable data, and clinical trial outcomes processes Alice's profile:

\begin{itemize}
    \item \textbf{Input features:} Demographics, comorbidities, prior medication history, sleep patterns from wearables, self-reported triggers, genetic markers (if available)
    \item \textbf{Output:} Ranked intervention candidates with uncertainty estimates, which in this scenario are the estimated probabilities that the treatment is optimal across all candidate treatments, which is based on the uncertainty in the estimated efficacy and the estimated efficacy, compared to the other treatments
\end{itemize}

\noindent The LFM may generate the following recommendations:

\begin{table}[h]
\centering
\small
\begin{tabular}{lccc}
\hline
\textbf{Intervention} & \textbf{Predicted} & \textbf{Probability} & \textbf{Trigger} \\
 & \textbf{Efficacy} & \textbf{being}  & \textbf{N-of-1?} \\
 &  & \textbf{optimal ($\sigma$)}  & \\
\hline
Magnesium supplement & 0.72 & 0.30 & Yes \\
Sleep regularity & 0.68 & 0.32 & Yes \\
Propranolol 40mg & 0.65 & 0.15 & No \\
Caffeine reduction & 0.61 & 0.23 & No \\
\hline
\end{tabular}
\caption{LFM-generated intervention candidates for illustrative patient Alice, with example numbers. $\sigma$ denotes the probability of being the best treatment across all candidate treatments, where an uncertainty threshold $\tau = 0.25$ triggers validation in N-of-1 trial.}
\label{tab:alice_recommendations}
\end{table}

\noindent Propranolol and caffeine reduction, with probabilities of being optimal, $\sigma$, below the pre-specified threshold of $\tau=0.25$, cannot be recommended directly based on population evidence. However, magnesium and sleep regularity exceed the uncertainty threshold, triggering N-of-1 validation.

\subsection{Step 2: N-of-1 Trial Design}

Alice enrolls in a sequential N-of-1 trial comparing magnesium supplementation versus sleep regularity and placebo:

\begin{itemize}
    \item \textbf{Design:} 6 periods $\times$ 2 weeks each, block-randomized, yielding e.g. (ABC)(BCA) or (BAC)(ABC) sequence
    \item \textbf{Primary outcome:} Migraine days per period (self-reported via mobile app)
    \item \textbf{Secondary outcomes:} Pain intensity (0--10 scale), functional disability, medication use
    \item \textbf{Data collection:} Daily headache diary, wearable-tracked sleep quality
\end{itemize}

\subsection{Step 3: Bayesian Updating and Results}

After 12 weeks, Alice's individual data are analyzed:

\begin{equation}
P(\theta_{\text{Alice}} | D_{\text{Alice}}) \propto P(D_{\text{Alice}} | \theta_{\text{Alice}}) \cdot P(\theta_{\text{Alice}} | \theta_{\text{pop}})
\end{equation}

\noindent where $\theta_{\text{pop}}$ represents the LFM's population prior and $D_{\text{Alice}}$ is Alice's trial data.

\noindent \textbf{Results:} A result may be that for Alice, the posterior probability that migraine days are reduced by at least 2 days per month, is 90\% while the same probability is 70\% when implementing a sleep regularity protocol.
Her personalized digital twin is updated accordingly, and the system may proceed to test further candidate interventions in subsequent trials if desired.

\subsection{Privacy Preservation Vignette}

Throughout this process, all trial data remain on Alice's device, only aggregated, anonymized effect estimates are shared (with consent) to improve population priors, and no raw health records leave the local environment.

\subsection{Privacy-Preserving Implementation}

A key advantage of the hybrid framework is that individual-level experimentation can occur locally, minimizing data exposure. Table~\ref{tab:privacy_framework} summarizes the privacy-preserving architecture across all framework components.

\begin{table*}[!htbp] 
\centering
\small
\begin{tabular}{|p{2.5cm}|p{2.4cm}|p{3.0cm}|p{2.5cm}|p{3.2cm}|}
\hline
\textbf{Component} & \textbf{Location} & \textbf{Data Handled} & \textbf{Privacy Mechanism} & \textbf{Guarantee} \\
\hline
\multicolumn{5}{|l|}{\textit{On-Device (Local) Components}} \\
\hline
Raw data storage & User device & Wearables, self-reports, EHR excerpts & Local encryption (AES-256) & Data never leaves device \\
\hline
Trial execution & User device & Randomization, outcome tracking & Fully local computation & Complete privacy \\
\hline
Posterior update & User device & Individual effect estimates & Bayesian update on-device & No transmission required \\
\hline
Digital twin & User device & Personalized model weights & Local fine-tuning only & User-controlled \\
\hline
\multicolumn{5}{|l|}{\textit{Server-Side Components}} \\
\hline
LFM inference & Server & Feature embeddings (not raw data) & Embedding projection & Reconstruction-resistant \\
\hline
Population priors & Server & Aggregated statistics only & Secure aggregation \cite{bonawitz_practical_2017} & Individual contributions hidden \\
\hline
Prior contribution & Server (opt-in) & Clipped gradients + noise & Differential privacy \cite{abadi2016deep} & $(\epsilon, \delta)$-DP \\
\hline
Trial templates & Server & Generic protocols & Public, non-sensitive & N/A \\
\hline
\end{tabular}
\caption{Privacy-preserving architecture of the hybrid framework. On-device components handle all sensitive personal data locally, while server components operate only on privacy-protected aggregates or non-sensitive templates.}
\label{tab:privacy_framework}
\end{table*}

\subsection{Safety, Regulatory, and Equity Considerations}

The deployment of hybrid LFM-N-of-1 systems in healthcare raises important safety, regulatory, and fairness concerns. Tables~\ref{tab:safety_framework} and~\ref{tab:equity_framework} summarize our approach to responsible implementation.

\begin{table*}[!htbp] 
\centering
\small
\begin{tabular}{|p{2.0cm}|p{3.0cm}|p{4.2cm}|p{4.8cm}|}
\hline
\textbf{Category} & \textbf{Concern} & \textbf{Mitigation Strategy} & \textbf{Implementation} \\
\hline
\multicolumn{4}{|l|}{\textit{Intervention Risk Stratification}} \\
\hline
\multicolumn{2}{|l|}{Tier 1: Low (Examples: lifestyle, supplements)} & Fully N-of-1 eligible if intervention can be tested & Informed consent + educational materials \\
\hline
\multicolumn{2}{|l|}{Tier 2: Medium (Example: common medications)} & N-of-1 with clinical oversight & Healthcare provider integration; IRB review for systematic deployment \\
\hline
\multicolumn{2}{|l|}{Tier 2: High (Examples: surgery, chemotherapy)} & LFM hypothesis only & RCT evidence required; no self-experimentation \\
\hline
\multicolumn{4}{|l|}{\textit{Adverse Event Monitoring}} \\
\hline
Safety signals & Unexpected symptoms & Automated stopping rules & Pre-defined severity thresholds trigger trial termination \\
\hline
Escalation & Serious events & Provider alerts & Real-time notification to healthcare team \\
\hline
Carryover effects & Inter-period contamination & Washout enforcement or statistical modeling & Mandatory washout periods between intervention phases or post-trial modeling \\
\hline
\multicolumn{4}{|l|}{\textit{LFM-Specific Safeguards}} \\
\hline
Hallucination & False recommendations & Retrieval augmentation & Ground outputs in verified clinical knowledge bases \\
\hline
Contraindications & Unsafe suggestions & Automated checking & Validate against patient-specific contraindication lists \\
\hline
Overconfidence & Miscalibrated certainty & Confidence thresholds & Flag low-reliability recommendations for clinician review \\
\hline
\multicolumn{4}{|l|}{\textit{Regulatory Compliance}} \\
\hline
SaMD classification & FDA/EU MDR scope & Intended use specification & Decision support vs. autonomous recommendation delineation \\
\hline
Continuous learning & Post-market updates & Surveillance protocols & Monitor performance drift; document model updates \\
\hline
Transparency & Algorithmic opacity & Documentation requirements & Architecture, training data, uncertainty methods disclosed \\
\hline
\end{tabular}
\caption{Safety and regulatory framework for hybrid LFM-N-of-1 systems. Interventions are stratified by risk tier, with corresponding oversight requirements and technical safeguards.}
\label{tab:safety_framework}
\end{table*}

\begin{table*}[!htbp] 
\centering
\small
\begin{tabular}{|p{2.7cm}|p{3.0cm}|p{4.0cm}|p{4.5cm}|}
\hline
\textbf{Category} & \textbf{Challenge} & \textbf{Mitigation Strategy} & \textbf{Implementation} \\
\hline
\multicolumn{4}{|l|}{\textit{Algorithmic Bias}} \\
\hline
Underrepresentation & Uneven error rates across groups & Stratified uncertainty monitoring & Elevated uncertainty for underrepresented groups triggers N-of-1 validation more readily \\
\hline
Biased proxies & Historical disparities encoded & Fairness-aware priors & Regularize to reduce reliance on biased outcome proxies \cite{obermeyer_dissecting_2019} \\
\hline
Validation gaps & Limited diverse testing & Inclusive trial design; summary data of trials fed back into population LFM & Multilingual interfaces; culturally appropriate outcome measures \\
\hline
\multicolumn{4}{|l|}{\textit{Digital Divide}} \\
\hline
Device access & No smartphone/wearable & Low-tech alternatives & Paper-based outcome tracking with periodic digitization \\
\hline
Digital literacy & Limited tech proficiency & Community health workers & Proxy data entry; in-person support for trial participation \\
\hline
Connectivity & Limited/no internet & Offline functionality & Local trial execution; opportunistic sync when connected \\
\hline
Cost barriers & Device/data expenses & Device lending programs & Healthcare system partnerships; subsidized connectivity \\
\hline
\multicolumn{4}{|l|}{\textit{Inclusive Design}} \\
\hline
Visual impairment & Screen-based interfaces & Screen reader compatibility & WCAG 2.1 AA compliance
; voice-based interaction \\
\hline
Motor impairment & Fine motor requirements & Simplified input methods & Large touch targets; voice input; switch access \\
\hline
Cognitive load & Complex trial protocols & Adaptive simplification & Step-by-step guidance; reminder systems; caregiver modes \\
\hline
Language barriers & English-only systems & Localization & Multilingual UI; culturally adapted content \\
\hline
\multicolumn{4}{|l|}{\textit{Governance}} \\
\hline
Benefit distribution & Efficiency gains to systems only & Patient-centered outcomes & Transparent reporting; patient advocates in governance \\
\hline
Accountability & Unclear responsibility & Governance structures & Clear liability frameworks; open-source components for verification \\
\hline
\end{tabular}
\caption{Equity and accessibility framework ensuring the hybrid system serves diverse populations. Addresses algorithmic fairness, digital access barriers, inclusive design principles, and governance structures.}
\label{tab:equity_framework}
\end{table*}

\section{Conclusion}

Large foundation models trained on population data face four structural tensions when providing personalized health interventions: personalization-external validity (optimizing within contexts degrades generalization across contexts), data-privacy (personalization requires comprehensive data yet privacy demands minimization), population-individual scale (models need massive populations yet must serve heterogeneous individuals), and algorithmic efficiency-human care (automation risks dehumanizing the listening, empathy, and trust-building essential to medicine). We propose resolving these tensions through a hybrid framework combining population-level foundation models with individual-level N-of-1 trials, in which adaptive digital twins inherit population priors, conduct systematic experiments on personal intervention strategies, and update via Bayesian inference based on individual outcomes. This approach quantifies uncertainty and triggers validation when needed, preserves privacy through local experimentation, reserves costly experiments for high-stakes scenarios where population knowledge is insufficient, and maintains patient agency through interpretable evidence generation. Beyond healthcare, this framework offers a principled path for any domain that requires personalized AI systems: balancing the power of large-scale learning with the rigor of individual-level causal inference to create AI that is not only intelligent at scale but also trustworthy, transparent, and genuinely personalized for each individual.

\bibliography{aaai2026}

\end{document}